\documentclass[runningheads]{llncs}
\usepackage[T1]{fontenc}
\usepackage{graphicx}
\usepackage{booktabs}
\usepackage[misc]{ifsym}
\usepackage[misc]{ifsym}
\usepackage{mwe}

\usepackage{amsmath, amssymb}
\usepackage{makecell}

\usepackage{multirow}

\usepackage{etoolbox}

\begin{document}

\title{NeuronTune: Fine-Grained Neuron Modulation for Balanced Safety-Utility Alignment in LLMs}

\titlerunning{NeuronTune}

\author{
Birong Pan\inst{1}\textsuperscript{*} \and
Jianhao Chen\inst{1,2}\textsuperscript{*} \and
Qiankun Pi \inst{1} \and
Mayi Xu \inst{1} \and
Yuanyuan Zhu \inst{1} \and
Ming Zhong\inst{1} \and
Tieyun Qian\inst{1,2} \textsuperscript{\Letter}
}

\authorrunning{B. Pan et al.}

\institute{
School of Computer Science, Wuhan University, Wuhan, China\\
\email{
panbirong@whu.edu.cn,
chenjianhao@whu.edu.cn,
qty@whu.edu.cn
}
\and
Zhongguancun Academy, Beijing, China\\
}

\maketitle              

\begingroup
\renewcommand{\thefootnote}{*}
\footnotetext{Birong Pan and Jianhao Chen contributed equally to this work.}
\endgroup

\begin{abstract}
Ensuring robust safety alignment while preserving utility is critical for the reliable deployment of Large Language Models (LLMs). However, current techniques fundamentally suffer from intertwined deficiencies: insufficient robustness against malicious attacks, frequent refusal of benign queries, degradation in generated text quality and general task performance, the former two reflecting tensions in \textbf{robust safety} and the latter constituting \textbf{utility impairment}. We attribute these limitations to the coarse-grained layer-wise interventions in existing methods. 
To resolve this, we propose \textit{NeuronTune}, a fine-grained framework that \textbf{pinpoints and modulates sparse neurons} to achieve simultaneous safety-utility optimization. Our approach first pinpoints safety-critical and utility-preserving neurons across all layers via attack-aware attribution, then adapts meta-learning to adaptively modulate their activations. Crucially, the intervention scope of NeuronTune is dynamically controlled via neuron inclusion thresholds, providing a flexible mechanism to prioritize either security-critical or utility-priority requirements. Extensive experimental results demonstrate that our method outperforms existing state-of-the-art technologies, achieving superior model safety while maintaining excellent utility.

\end{abstract}

\section{Introduction}

Large Language Models (LLMs) \cite{hurst2024gpt,2023llama} demonstrate remarkable capabilities across diverse tasks, yet remain highly vulnerable to malicious attacks such as jailbreaking \cite{jin2025guard,xiao-etal-2024-distract}, which can induce harmful or uncontrolled outputs. Ensuring robust safety alignment with human values is a critical prerequisite for their secure and reliable deployment. 
Traditional safety alignment methods, such as full model fine-tuning and complex Reinforcement Learning from Human Feedback (RLHF), face challenges related to prohibitive computational costs or inherent instability. Therefore, lightweight techniques are highly attractive due to their efficiency and low computational cost. However, such techniques \cite{zhang2025controllablesafetyalignmentinferencetime} face fundamental challenges in simultaneously achieving safety robustness and utility preservation \cite{rottger-etal-2024-xstest,safetyandoverdefensiveness}. The former dilemma manifests as either inadequate resistance to malicious attacks or exaggerated stringency in rejecting benign queries. The latter presents as the generation of low-quality responses. Both issues severely limit practical applicability. 

We contend that these limitations stem from the coarse-grained intervention strategies. Such approaches typically identify safety-critical layers using techniques like hidden-state comparisons or steering vectors \cite{Cao2024SCANSMT,wangdetoxifying}, and then apply uniform modulation across all layers. 
However, coarse-grained adjustments fail to precisely pinpoint the key factors related to safety and utility. Furthermore, uniform layer-wise interventions lack the necessary flexibility to accommodate the diverse adjustment requirements of different influencing factors, leading to either insufficient defense or excessive over-refusal.
Consequently, the inherent safety-utility trade-off remains unresolved.
To effectively mitigate these issues, it becomes imperative to precisely identify and modulate significant factors.

Therefore, we propose a fine-grained neuron-level safety alignment framework named \textbf{NeuronTune}. 
Specifically, we first pinpoint the neurons impacting the safety-utility balance. Drawing inspiration from the knowledge neurons \cite{dai-etal-2022-knowledge}, we assume that safety-critical features and utility-preserving knowledge are intrinsically stored within specific neurons. 
To effectively identify these neurons, we propose \textit{an attack-aware attribution} method, which traces safety-crucial and utility-related neurons by performing attribution on safe and useful responses when harmful or benign queries are subjected to adversarial perturbations.

Secondly, we edit the pinpointed neurons via adaptive activation adjustment. Building on the insights that meta-learning \cite{Finn2017ModelAgnosticMF} enables superior performance across diverse tasks, we reveal that this co-optimization framework facilitates the modulation of distinct types of neurons. It effectively circumvents the inherent limitations of uniform regulation. 
Unlike the standard meta-learning which often modifies entire model parameters for new tasks, we propose to adapt it to optimize the scaling factors of pinpointed neurons. This fine-grained control is crucial for avoiding the pitfalls of coarse-grained interventions.
Moreover, we design a dynamic control mechanism, which allows users to flexibly tune the intervention scope, such as tightening safety by modulating more safe neurons in high-risk scenarios, or prioritizing utility via conservative neuron selection. This configurability provides a flexible adjustment for diverse deployment needs.

The primary contributions are as follows.
\begin{enumerate}
    \item \textit{Method Innovation}: We introduce a fine-grained, neuron-level alignment framework that integrates neuron localization based on attack-aware attribution with adaptive neuron modulation guided by meta-learning, striking a delicate balance between safety and utility.
    \item \textit{Tunable Mechanism}: Our design introduces a tunable intervention system, which facilitates model adaptation to diverse safety and utility demands through the regulation of neuron counts.
    \item \textit{Empirical Validation}: Extensive experiments on multiple benchmarks and LLMs demonstrate superior effectiveness and adaptability over baselines.
\end{enumerate}

\section{Related Work}

\paragraph{Safety Alignment} 
A considerable body of research has been devoted to enhance models safety, broadly termed safety alignment. These works mainly focus on the model alignment through techniques such as supervised fine-tuning \cite{Bianchi2023SafetyTunedLL} or training-free interventions \cite{cao-etal-2024-defending}. 
However, the majority of existing research primarily focuses on improving the safety performance of LLMs. While crucial, an exclusive focus on safety enhancement often inadvertently leads to exaggerated safety \cite{rottger-etal-2024-xstest} and pretrained knowledge degradation \cite{lin-etal-2024-mitigating}. Despite attempts through manipulating representation to mitigate these limitations \cite{Cao2024SCANSMT}, these techniques operate at a broad level of granularity, failing to strike a good balance between robust safety and utility preservation.

\paragraph{Knowledge Neurons} The concept of knowledge neurons \cite{dai-etal-2022-knowledge} has been proposed as a way to interpret the behaviors of models by modifying specific neurons, thereby influencing the model’s generation. Neuron-level pruning methods have been developed to identify task-critical neurons. For instance, IRCAN \cite{IRCAN} calculates the importance scores of all neurons based on their contribution to the loss, while Wanda \cite{sun2024simpleeffectivepruningapproach} tracks changes in the immediate outputs of each layer when specific neurons are pruned. Regarding safety neurons, \cite{chen2024findingsafetyneuronslarge} introduces generation-time activation contrasting to locate safety neurons, highlighting their sparse distribution. 
\cite{10.5555/3692070.3694226,Wu2025NeuroStrikeNA} use simple neuron removal to identify important neurons.
Building on these insights, our approach focuses on balancing safety and utility by identifying and adaptively regulating the neurons associated with both.

\section{Preliminaries}
In this section, we begin by examining the limitations of coarse-grained layer-wise interventions on the problems of exaggerated safety and utility degradation. 

Existing safety alignment methods often resort to coarse-grained interventions, like modifying entire layers. While these approaches aim to enhance safety, they frequently introduce undesirable side effects, notably exaggerated safety and utility degradation.

To empirically investigate the impact of such coarse-grained interventions, we conduct an analysis using the DINM baseline method \cite{wangdetoxifying} on LLaMA-3.1-8B-Instruct. We select DINM for this empirical investigation since it is a representative method that employs layer-wise interventions to mitigate unsafe behaviors. Moreover, DINM allows to vary the percentage of parameters updated within a layer, which enables us to systematically observe the direct consequences of such interventions on both safety and utility. 
Table \ref{tab: coarse-grained analysis} shows the results by employing DINM to edit the varying percentages of model parameters within a layer.

As shown in Table \ref{tab: coarse-grained analysis}, when a large percentage of parameters are updated, the model achieves sufficient safety. However, this comes at a severe cost, i.e., significant exaggerated safety and drastic utility degradation. This indicates that the model becomes overly cautious, refusing even harmless requests and generating low-quality, repetitive, or uninformative text.
A gradual mitigation of exaggerated safety and a recovery in general utility are observed as the percentage of updated parameters decreases.
However, it is accompanied by a decline in sufficient safety, e.g., SafeEdit Refusal Rate decreases, implying that the model becomes less robust against harmful queries. This analysis clearly demonstrates that coarse-grained, layer-wise interventions struggle to simultaneously achieve robust safety and maintain high utility, often forcing a compromise between these two crucial aspects. 

The aforementioned drawback of coarse grained, layer-wise interventions underscores the necessity for a more nuanced and fine-grained approach.

\begin{table}[ht]
\centering
\caption{Results of employing DINM to edit varying percentages of parameters within a layer on LLaMA3.1-8B-Instruct. $\uparrow$ indicates that higher values are better, while $\downarrow$ indicates that lower values are better.}  
\label{tab: coarse-grained analysis}
\begin{tabular}{lccccc}
\toprule
\multirow{3}{*}{\textbf{Methods}} & \multicolumn{2}{c}{\textbf{SafeEdit}} & \multicolumn{2}{c}{\textbf{Alpaca}} & \multicolumn{1}{c}{\textbf{MMLU}} \\
\cmidrule(lr){2-3} \cmidrule(lr){4-5} \cmidrule(lr){6-6}
& \textbf{Refuse Rate$\uparrow$} & \textbf{Entropy$\uparrow$} & \textbf{Refuse Rate$\downarrow$} & \textbf{Entropy$\uparrow$} & \textbf{Accuracy$\uparrow$} \\
\midrule
DINM (100\%) & 100.00\% & 1.132 bit & 98\% & 1.089 bit & 0.00\% \\
DINM (50\%) & 100.00\% & 1.142 bit & 100\% & 1.233 bit & 0.01\% \\
DINM (20\%) & 99.49\% & 1.616 bit & 86\% & 1.778 bit & 60.00\% \\
DINM (10\%) & 82.46\% & 2.545 bit & 42\% & 3.760 bit & 59.64\% \\
DINM (5\%) & 42.67\% & 5.019 bit & 4\% & 5.520 bit & 61.97\% \\
\bottomrule
\end{tabular}
\end{table}

\section{Method: NeuronTune}
\begin{figure}[h]
\centering
\includegraphics[width=0.8\columnwidth]{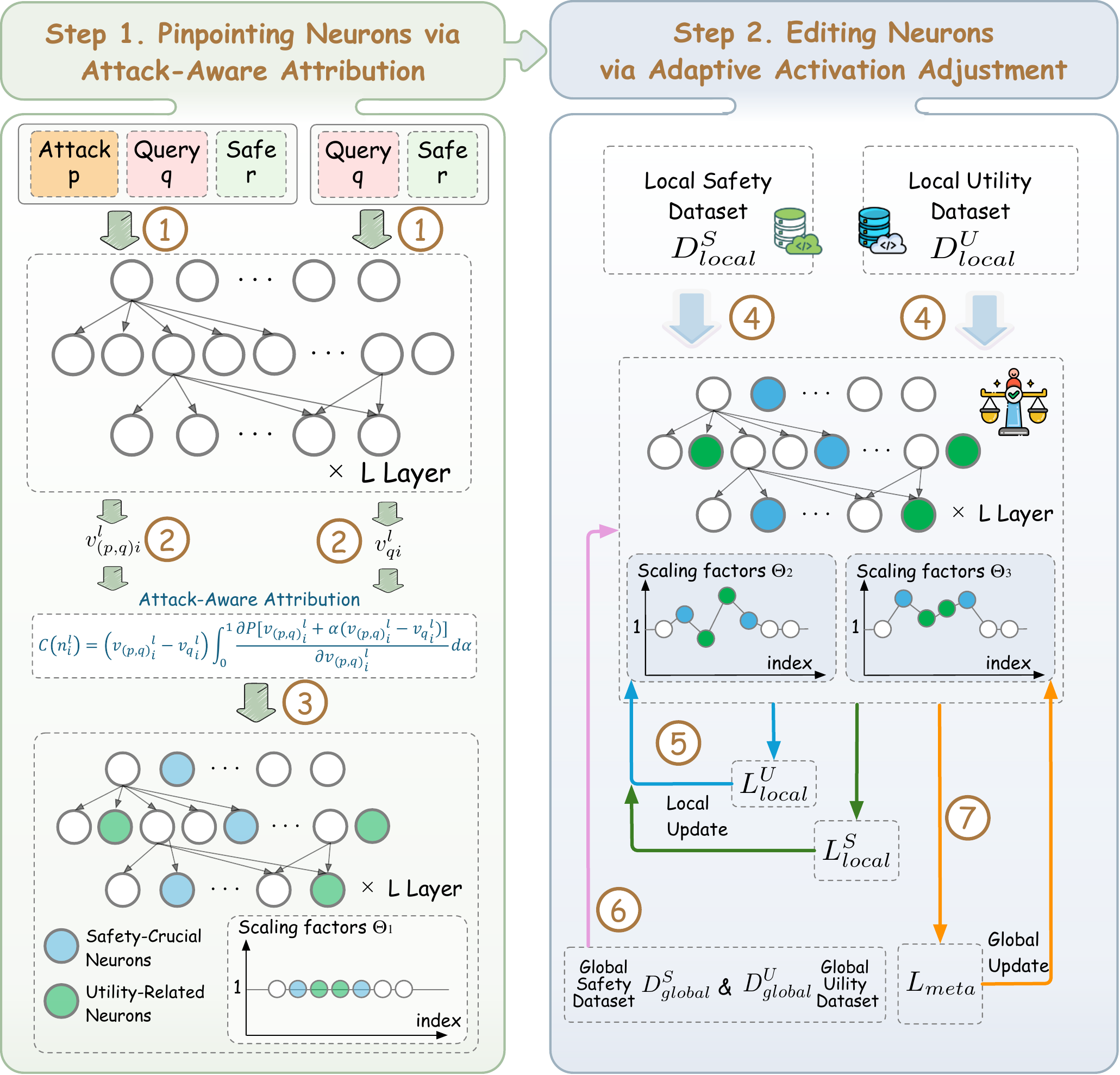}
\caption{The overview of NeuronTune, containing pinpointing neurons and adaptively modulating neurons. The numbered steps represent the sequential order of processing.}
\label{fig:framework}
\end{figure}

Building upon the insights regarding the limitations of coarse-grained interventions, and recognizing that model capabilities are encoded within individual neurons \cite{dai-etal-2022-knowledge}, we propose NeuronTune, a novel fine-grained method for safety-utility alignment. 
Our approach addresses the imperative to achieve robust safety and mitigate utility degradation, Fig.~\ref{fig:framework} provides an overview of NeuronTune. NeuronTune consists of two primary stages. First, \textit{Pinpointing Safety and Utility Neurons via Attack-Aware Attribution}, where we identify safety-crucial and utility-related neurons. Second, \textit{Editing Neurons via Adaptive Activation Adjustment}, where we dynamically adjust these neurons.

\subsection{Pinpointing Safety and Utility Neurons via Attack-Aware Attribution}
Following prior neuron-level analysis, we define a neuron as a single dimension of the MLP intermediate activation before the down projection:
\(z=\sigma(W_{\mathrm{gate}}x)\odot W_{\mathrm{up}}x\).
While LLMs effectively handle direct harmful queries, their robustness significantly degrades against adversarial attacks. Since safety-related knowledge of model is intrinsically encoded in neural activations, these attacks expose vulnerabilities in specific safety-critical neurons.
To capitalize on this, we identify these key neurons by performing attribution on safe responses generated even when the model is subjected to adversarial harmful queries. Simultaneously, analyzing useful responses to benign queries under similar adversarial contexts allows us to isolate neurons responsible for utility preservation.
Precisely pinpointing these two functional neuron sets, those safeguarding the model and those sustaining utility, is crucial for achieving a better balance between robust safety and utility.
Empirically, adversarial prompts substantially reduce harmful-query refusal rates, e.g., from 47.49\% to 10.77\% on LLaMA3.1-8B-Instruct and from 82.67\% to 14.97\% on Qwen2.5-14B-Instruct, motivating attack-aware neuron attribution.

To this end, we propose an attack-aware attribution method designed to pinpoint neurons whose activations are vital for processing adversarial contexts and ensuring desired safety-compliant and functionally useful outputs.

For safety-critical neurons, given an input pair $(p,q)$, where $p$ represents the misleading adversarial prompt and $q$ represents the harmful question, we aim to guide the model towards a desired safe response, denoted as $r_{a}$. 
Our approach calculates the contribution score of each neuron in perceiving the prompt's influence on the generation of the safe response. 
Initially, we take only $q$ as input, record the activation value of each neuron and denote it as ${\boldsymbol{v}_{q}}_i^l$. Subsequently, we input both $p$ and $q$ into the model and record the new activation value, denoted as ${\boldsymbol{v}_{(p,q)}}_i^l$. To calculate the contribution score $\mathrm{C}(n_i^l)$, we gradually change the activation value of a neuron $n_i^l$ from ${\boldsymbol{v}_{q}}_i^l$ to ${\boldsymbol{v}_{(p,q)}}_i^l$. At the same time, we calculate the probability of the answer $r_{a}$ predicted by model, denoted as: 
\begin{equation}
    P(\boldsymbol{v}_i^l)=p(r_{a}|p,q,\mathbf{A}(n_i^l)=\boldsymbol{v}_i^l) ,
\end{equation}
where $\boldsymbol{v}_i^l$ is a given value assigned to the neuron activation $\mathbf{A}(n_i^l)$. We integrate the gradient of the probability during this process as the neuron's contribution score, as follows:
\begin{equation}
\scalebox{0.9}{$\mathrm{C}(n_i^l) = \left({\boldsymbol{v}_{(p,q)}}_i^l-{\boldsymbol{v}_{q}}_i^l\right) \int_{\alpha=0}^{1} \frac{\partial P\left[{\boldsymbol{v}_{q}}_i^l + \alpha \left({\boldsymbol{v}_{(p,q)}}_i^l-{\boldsymbol{v}_{q}}_i^l\right)\right]}{\partial {\boldsymbol{v}_{(p,q)}}_i^l} \, d\alpha$},
\end{equation}
where $ \frac{\partial P\left[{\boldsymbol{v}_{q}}_i^l + \alpha \left({\boldsymbol{v}_{(p,q)}}_i^l-{\boldsymbol{v}_{q}}_i^l\right)\right]}{\partial {\boldsymbol{v}_{(p,q)}}_i^l} $ calculates the gradient of the model probability with regard to $ {\boldsymbol{v}_{(p,q)}}_i^l $, $\alpha$ controls the integration from ${\boldsymbol{v}_{q}}_i^l$ to ${\boldsymbol{v}_{(p,q)}}_i^l$.

By integrating over the gradient as $\alpha$ changes from 0 to 1, $\mathrm{C}(n_i^l)$ accumulates the output probability changes caused by the activation value changes from the absence to the presence of adversarial attacks. If the neuron has a strong perception and processing capability regarding the attack, the gradient will be significant, resulting in a large integration value. Therefore, the attribution score can measure the neuron’s sensitivity to the context and its contribution to processing the context.
We then select a subset of top-k neurons with the highest score as the safety-crucial neurons, $\mathcal{N}_s$.

Similarly, for a set of benign, general knowledge questions, we calculate the gradient of the log-likelihood of high-quality responses with respect to neuron activations. Neurons with high contribution scores in this context are identified as utility-related $\mathcal{N}_u$, as they are essential for the general performance and ability to generate informative text. 

\subsection{Editing Neurons via Adaptive Activation Adjustment}
Having identified the safety-crucial and utility-related neurons, the next pivotal step in NeuronTune is to edit these neurons to achieve a harmonious balance between robust safety and preserved utility.
The magnitude of suppressing or amplifying specific neuron activations notably affects the expression of the corresponding knowledge. Distinct regulatory intensities are required for the two types of neurons given their functional divergence. Therefore, precise modulation of neurons is essential.
Capitalizing on the synergistic optimization capabilities of MAML \cite{Finn2017ModelAgnosticMF}, we apply it to refine the scaling factors $\Theta$ for both categories of neurons. 
Unlike standard $\text{MAML}$ which optimizes entire model parameters, NeuronTune focuses on modulating the scaling factors. This makes the optimization lightweight and fast, as the core parameters remain locked.

The modulation process is initialized with prior guidance: safety-critical and utility-related neurons are assigned scaling factors to steer the model toward safer outputs while mitigating utility degradation.
The optimization is an iterative bi-level process. At the start of the inner loop, the current factors $\Theta$ are cloned to create a temporary set $\Theta'$. This cloning is crucial to maintain $\Theta$'s status as the global initial point. $\Theta'$ is then rapidly updated using local safety and utility data ($\mathcal{D}_{\text{local}}^{S}$, $\mathcal{D}_{\text{local}}^{U}$): 
\begin{equation}
    \Theta' \leftarrow \Theta' - \eta_{\text{inner}} \nabla_{\Theta'} \mathcal{L}_{\text{local}}
\end{equation}
The local loss $\mathcal{L}_\text{local}$, which consists of the safety-related component $\mathcal{L}_{\text{local}}^{S}$ and utility-related component $\mathcal{L}_{\text{local}}^{U}$, is formally defined as the conditional cross-entropy loss over a sequence of targeted tokens.

The outer loop evaluates the quality of the original initial point $\Theta$ by examining the performance of the adapted set $\Theta'$ on global data ($\mathcal{D}_{\text{global}}^{S}$, $\mathcal{D}_{\text{global}}^{U}$).
The meta loss $\mathcal{L}_{\text{meta}}$ is defined as the joint loss of the global safety loss $\mathcal{L}_{\text{global}}^{S}$ and the global utility loss $\mathcal{L}_{\text{global}}^{U}$:
\begin{equation}
    \mathcal{L}_{\text{meta}} = \lambda \mathcal{L}_{\text{global}}^{S}(\Theta') + (1 - \lambda) \mathcal{L}_{\text{global}}^{U}(\Theta')
\end{equation}
Then $\Theta$ is optimized to be the best possible initialization point, guaranteeing the learned modulation strategy generalizes effectively to strike the optimal safety-utility balance:
\begin{equation}
    \Theta \leftarrow \Theta - \eta_{\text{meta}} \nabla_{\Theta} \mathcal{L}_{\text{meta}}(\Theta')
\end{equation}
Through this mechanism, our approach precisely tunes the influence of related neurons.

Furthermore, to accommodate diverse practical application scenarios and varying demands for safety or utility, NeuronTune features a tunable mechanism. This mechanism allows for the flexible selection of the number of neurons to be regulated. By controlling the quantity of modulated neurons, users can fine-tune the model's behavior, emphasizing either heightened safety, achieved by regulating more safety-critical neurons, or enhanced utility, by prioritizing utility-related neurons, thereby adapting to specific deployment requirements.

\section{Experiment}

\subsection{Experimental Setup}
\paragraph{Models} We select four representative LLMs: LLaMA2-7B-Chat, LLaMA3.1-8B-Instruct, Qwen2.5-7B-Instruct and Qwen2.5-14B-Instruct, to evaluate the effectiveness and scalability of NeuronTune in balancing safety and utility.

\paragraph{Safety and Utility Evaluation Datasets}
We evaluate model performance from two critical perspectives: robust safety and utility. The former encompasses sufficient safety and exaggerated safety.
For sufficient safety, which measures the ability to resist generating harmful content, we utilize SafeEdit \cite{wangdetoxifying} and AdvBench \cite{zou2023universaltransferableadversarialattacks}, two widely recognized safety benchmarks. 

For meta-optimization, \(D_{local}\) and \(D_{global}\) are constructed by proportional random sampling from the original datasets, with \(D_{local}\cap D_{global}=\emptyset\) to avoid data leakage.
For exaggerated safety, we consider two prominent benchmarks: Alpaca \cite{alpaca} and TruthfulQA \cite{lintruthfulqa}. Specifically, we leverage the benign queries within these datasets. Our evaluation quantifies the rate at which models incorrectly refuse to answer these benign inputs.

\paragraph{Baselines} We compare NeuronTune with the following baselines. (1) DINM \cite{wangdetoxifying} is a editing approach to identify toxic regions and mitigate unsafe behaviors in LLMs. (2) SCANS \cite{Cao2024SCANSMT} applies refusal steering vectors to identify and mitigate such exaggerated safety behaviors in LLMs. (3) CAVGAN \cite{li2025cavganunifyingjailbreakdefense} leverages the security judgment boundary to achieve efficient defense.

\paragraph{Metrics}
For both sufficient safety and exaggerated safety, we employ the Refusal Rate, defined as the proportion of queries (whether harmful or benign) that are rejected. 
The determination of whether a query should be rejected is used a classifier \cite{wangdetoxifying}, which evaluates the safety of response content and rejects those deemed unsafe. 
For utility, we assess three complementary dimensions. (1) Information Content: Measured by entropy of generated text, reflecting the diversity and richness of the output. (2) Fluency: Evaluated using perplexity-based metrics to assess the linguistic smoothness and naturalness of responses generated by the aligned LLM. (3) General Capability: Evaluated on MMLU \cite{Hendrycks2020MeasuringMM}, selected for its comprehensive coverage of knowledge-intensive tasks across various domains.

Robust safety performance is calculated by the Refusal Rate for harmful and benign queries. Utility is primarily represented by the ENTropy of the generated text for both harmful and benign queries. Given that these performance indicators have different units and value ranges, we first normalize both metrics, recorded as $RR_{norm}$ and $ENT_{norm}$, and then adopt an F1-score-like calculation:
\begin{equation}
    SU-F1= \frac{2 \times RR_{norm} \times ENT_{norm}}{RR_{norm} + ENT_{norm}}
\end{equation}
This combined metric, Safety-Utility F1 (SU-F1), evaluates the overall balance between robust safety and utility.

We compute the normalized refusal and entropy scores as
\[
RR_{norm}=\frac{RR_{unsafe}+(1-RR_{safe})}{2},
\]
\[
ENT_{norm}=\frac{ENT_{unsafe}+ENT_{safe}-ENT_{min}}{ENT_{max}-ENT_{min}},
\]
where \(ENT_{min}\) and \(ENT_{max}\) are determined over the entire evaluation suite.

\subsection{Main Results}

\paragraph{Performance in terms of safety-utility balance}
As shown in Table \ref{tab:main}, baseline methods either lack sufficient capability to resist adversarial attacks, or they exhibit excessive rejection of benign queries alongside degraded response quality. Our NeuronTune consistently achieves the highest SU-F1 score across all evaluated LLMs. 

In summary, the experimental results clearly demonstrate that NeuronTune consistently achieves a superior balance between robust safety and preserved utility. By precisely modulating individual neurons, our method effectively enhances sufficient safety while simultaneously mitigating exaggerated safety and maintaining high general performance, thereby outperforming existing coarse-grained intervention methods and addressing the intertwined deficiencies.
\begin{table*}[ht]
\centering
\caption{Main results on safety and utility evaluation across different LLMs. \textbf{Bold} denotes the best performance in each block. }
\label{tab:main}
\resizebox{\linewidth}{!}{
\begin{tabular}{clccccc}
\toprule
\multirow{3}{*}{\textbf{Models}} & \multirow{3}{*}{\textbf{Methods}} & \multicolumn{2}{c}{\textbf{SafeEdit}} & \multicolumn{2}{c}{\textbf{Alpaca}}  & \multirow{3}{*}{\textbf{SU-F1$\uparrow$}} \\
\cmidrule(lr){3-4}
\cmidrule(lr){5-6}
 & & \textbf{Refuse Rate$\downarrow$} & \textbf{Entropy$\uparrow$} & \textbf{Refuse Rate$\downarrow$} & \textbf{Entropy$\uparrow$} \\
\midrule
\multirow{5}{*}{LLaMA2-7B-Chat} 
& Default & 38.56\% & 5.482 bit & 1\% & 5.736 bit & 0.698 \\
& DINM & 90.97\% & 4.125 bit & 41\% & 4.799 bit & 0.623 \\
& SCANS & 97.85\% & 3.038 bit & 1\% & 3.718 bit & 0.534 \\
& CAVGAN & 84.41\% & 5.213 bit & 20\% & 5.705 bit & 0.748 \\
& \textbf{NeuronTune} & 91.59\% & 5.014 bit & 1\% & 5.143 bit & \textbf{0.770} \\
\midrule
\multirow{5}{*}{LLaMA3.1-8B-Instruct} 
& Default & 10.77\% & 6.832 bit & 2\% & 6.077 bit & 0.660 \\
& DINM & 42.67\% & 5.019 bit & 4\% & 5.520 bit & 0.675 \\
& SCANS & 99.28\% & 1.490 bit & 0\% & 1.401 bit & 0.128 \\
& CAVGAN & 30.67\% & 6.388 bit & 3\% & 6.189 bit & 0.715 \\
& \textbf{NeuronTune} & 49.74\% & 5.317 bit & 3\% & 5.912 bit & \textbf{0.722} \\
\midrule
\multirow{5}{*}{Qwen2.5-7B-Instruct} 
& Default & 14.87\% & 6.518 bit & 3\% & 6.345 bit & 0.670 \\
& DINM & 91.79\% & 6.187 bit & 57\% & 6.208 bit & 0.731 \\
& SCANS & 35.90\% & 4.810 bit & 0\% & 4.946 bit & 0.635 \\
& CAVGAN & 40.82\% & 6.787 bit & 13\% & 6.620 bit & 0.740 \\
& \textbf{NeuronTune} & 53.13\% & 6.067 bit & 2\% & 6.096 bit & \textbf{0.768} \\
\midrule
\multirow{5}{*}{Qwen2.5-14B-Instruct} 
& Default & 14.97\% & 6.704 bit & 2\% & 6.608 bit & 0.685 \\
& DINM & 45.95\% & 6.734 bit & 6\% & 6.846 bit & 0.784 \\
& SCANS & 45.33\% & 6.130 bit & 3\% & 6.602 bit & 0.764 \\
& CAVGAN & 58.77\% & 6.894 bit & 11\% & 6.706 bit & 0.808 \\
& \textbf{NeuronTune} & 65.03\% & 6.592 bit & 4\% & 6.379 bit & \textbf{0.824} \\
\bottomrule
\end{tabular}
}
\end{table*}

\paragraph{Performance in terms of  more utility metrics} 
In our primary experiments, utility is assessed using the entropy of generated text. To more comprehensively evaluate the utility of the aligned model, we additionally assess accuracy on MMLU and the fluency of generated text.

The results for all methods are shown in Table \ref{tab: general performance}. While NeuronTune's MMLU accuracy is not the highest compared to DINM, it is crucial to consider the overall context of safety alignment. Our method significantly mitigates the exaggerated safety issue, which often comes at a substantial cost to utility in other approaches. Therefore, despite being suboptimal on MMLU alone, NeuronTune's ability to maintain a strong balance between robust safety and utility preservation demonstrates its overall effectiveness and practical advantage.
\begin{table}[htbp!]
\centering
\caption{Results for more utility metrics of LLaMA-3.1-8B-Instruct, including general capability, response fluency and entropy. The best results are in \textbf{bold} and the second best ones are in \underline{underlined}.}  
\label{tab: general performance}
\begin{tabular}{lllll}
\toprule
\multirow{1}{*}{\textbf{Methods}} & 
\multirow{1}{*}{\textbf{Accuracy$\uparrow$}} & 
\multirow{1}{*}{\textbf{Fluency$\uparrow$}} & 
\multirow{1}{*}{\textbf{Entropy$\uparrow$}} & \multirow{1}{*}{\textbf{Avg.$\uparrow$}} \\
\midrule
Defaults & 64.43 & 8.31 & 12.909 & 28.55 \\
DINM & 61.97 & 5.96 & 10.539 & \textbf{26.16} \\
SCANS & 38.83 & 1.45 & 2.891 & 14.39 \\
CAVGAN & 36.13 & 7.27 & 12.588 & 18.66 \\
NeuronTune & 60.86 & 6.34 & 11.229 & \underline{26.14} \\
\bottomrule
\end{tabular}
\end{table}

On the additional AdvBench and TruthfulQA evaluation, NeuronTune achieves the best overall SU-F1 of 0.822 on LLaMA3.1-8B-Instruct, slightly outperforming Default, DINM, SCANS, and CAVGAN with SU-F1 scores of 0.818, 0.706, 0.517, and 0.820, respectively, further supporting its safety-utility balance.
On the out-of-distribution MedSafetyBench~\cite{han2024medsafetybenchevaluatingimprovingmedical} evaluation, NeuronTune obtains the highest SU-F1 of 0.84, compared with 0.83, 0.74, 0.36, and 0.79 for Default, DINM, SCANS, and CAVGAN, indicating robustness to unseen adversarial safety scenarios.

\subsection{Ablation Study}
To validate the effectiveness of each component, we conduct an ablation study of NeuronTune when removing attribution-based neuron identification (w/o Pinpointing), and meta-learning-driven adaptive adjustment (w/o Adjustment) respectively.
The results in Table \ref{tab: ablation} confirm that each component of NeuronTune is indispensable for achieving its safety-utility balance. 
Removing any of these components leads to a significant degradation in overall performance, failing to find the optimal trade-off. This validates the synergistic design of NeuronTune.

Removing any of these components leads to a significant degradation in overall performance, either by compromising robust safety, inducing severe utility degradation, or failing to find the optimal trade-off. This validates the synergistic design of NeuronTune.
\begin{table}[ht]
\centering
\caption{Ablation study for NeuronTune on LLaMA3.1-8B-Instruct. w/o Pinpointing, w/o Adjustment removes attribution-based neuron identification, and meta-learning-driven adaptive adjustment, respectively.}  
\label{tab: ablation}
\begin{tabular}{lccccc}
\toprule
\multirow{3}{*}{\textbf{Methods}} & \multicolumn{2}{c}{\textbf{SafeEdit}} & \multicolumn{2}{c}{\textbf{Alpaca}} & \multirow{3}{*}{\textbf{SU-F1}$\uparrow$} \\
\cmidrule(lr){2-3} \cmidrule(lr){4-5}
&  \textbf{Refuse Rate$\downarrow$} & \textbf{Entropy$\uparrow$} & \textbf{Refuse Rate$\downarrow$} & \textbf{Entropy$\uparrow$} \\
\midrule
\textbf{NeuronTune} & 49.74\% & 5.317 bit & 3\% & 5.912 bit & \textbf{0.722} \\
w/o Pinpointing & 8.51\% & 6.556 bit & 1\% & 6.194 bit & 0.652 \\
w/o Adjustment & 93.74\% & 1.667 bit & 0\% & 2.109 bit & 0.239 \\
\bottomrule
\end{tabular}
\end{table}

\subsection{Adaptability Across Diverse Scenarios via Neuron Number Regulation}
This section details the experimental evaluation of NeuronTune's tunable mechanism, which facilitates model adaptation to diverse safety and utility demands through the regulation of neuron counts. Our approach allows for independent control over the number of safety-crucial and utility-related neurons. Specifically, after obtaining the contribution scores for each neuron via gradient attribution, we sort them by score in descending order and select the top-k neurons for modulation.
To assess the impact of varying neuron counts, we conduct experiments by dynamically adjusting the number of safety-crucial and utility-related neurons to 500, 1000, 1500, and 2000 based on a balance of empirical observation. 
From the results in Table \ref{tab:tunable_mechanism_results}, we observe two key phenomena.

\textit{Impact of Increasing Safety-Crucial Neurons while Keeping Utility Neurons Fixed} When the number of utility-related neurons is fixed, and the number of safety-crucial neurons is progressively increased, the model's performance on the adversarial attack dataset significantly improves, indicating enhanced sufficient safety. For instance, with 500 utility neurons, increasing safety neurons from 500 to 2000 boosts SafeEdit Refuse Rate from 41.33\% to 83.69\%. 
However, simply maximizing the number of safety neurons is not always optimal. Adjusting an excessive number of safety-crucial neurons can lead to over-safety issues and a degradation in text quality. This phenomenon is likely attributed to the existence of overlapping neurons. Therefore, it is essential to flexibly calibrate the neuron counts to achieve a balance tailored to specific application scenarios.

\textit{Impact of Increasing Utility-Related Neurons while Keeping Safety Neurons Fixed} When the number of safety-crucial neurons is fixed, and the number of utility-related neurons is progressively increased, sufficient safety also enhances, albeit to a lesser extent compared to increasing safety-crucial neurons. More importantly, the issues of exaggerated safety and degradation in text quality exhibit fluctuations. Even when a decrease is observed, its magnitude is generally smaller than that observed when increasing safety-crucial neurons with fixed utility-related ones.
\begin{table}[ht]
\centering
\caption{Results on dynamically adjusting the number of safety-crucial and utility-related neurons. The tunable mechanism offers a practical approach to adapt NeuronTune to different application scenarios.}
\label{tab:tunable_mechanism_results}
\resizebox{\linewidth}{!}{
\begin{tabular}{clccccc}
\toprule 
\multirow{2.5}{*}{\parbox{5cm}{\centering \textbf{Number of Safety-Crucial \\ / Utility-Related Neurons}}} & \multicolumn{2}{c}{\textbf{SafeEdit}} & \multicolumn{2}{c}{\textbf{Alpaca}} \\
\cmidrule(lr){2-3}
\cmidrule(lr){4-5}
 & \textbf{Refuse Rate$\uparrow$} & \textbf{Entropy$\uparrow$} & \textbf{Refuse Rate$\downarrow$} & \textbf{Entropy$\uparrow$} \\
\midrule
\multirow{1}{*}{500/500} 
& 41.33\% & 5.297 bit & 1\% & 6.077 bit \\
\multirow{1}{*}{1000/500} 
& 62.26\% & 4.245 bit & 0\% & 4.870 bit \\
\multirow{1}{*}{1500/500} 
& 74.05\% & 3.638 bit & 6\% & 4.267 bit \\
\multirow{1}{*}{2000/500} 
& 83.69\% & 3.285 bit & 7\% & 4.262 bit \\
\midrule
\multirow{1}{*}{500/1000} 
& 43.38\% & 5.483 bit & 3\% & 5.867 bit \\
\multirow{1}{*}{1000/1000} 
& 69.64\% & 4.377 bit & 0\% & 4.446 bit \\
\multirow{1}{*}{1500/1000} 
& 80.41\% & 3.771 bit & 1\% & 4.228 bit \\
\multirow{1}{*}{2000/1000} 
& 86.56\% & 3.270 bit & 9\% & 3.889 bit \\
\midrule
\multirow{1}{*}{500/1500} 
& 49.74\% & 5.317 bit & 3\% & 5.912 bit \\
\multirow{1}{*}{1000/1500} 
& 60.41\% & 4.217 bit & 1\% & 4.803 bit \\
\multirow{1}{*}{1500/1500} 
& 69.44\% & 4.029 bit & 1\% & 4.597 bit \\
\multirow{1}{*}{2000/1500} 
& 77.74\% & 3.568 bit & 8\% & 4.266 bit \\
\bottomrule
\end{tabular}
}
\end{table}

Overall, there indeed exist neurons that are strongly correlated with either safety or utility. Therefore, increasing the count of safety-related neurons directly enhances sufficient safety, while increasing utility-related neurons helps curb exaggerated safety and mitigate utility degradation. The tunable mechanism offers a practical approach to adapt NeuronTune to different application scenarios. For instance, in high-security demand scenarios, one can choose to modulate a larger number of safety-crucial neurons to prioritize robust defense. Conversely, in applications where maintaining high conversational quality and helpfulness is paramount, a greater emphasis can be placed on modulating utility-related neurons.

\subsection{Analysis of Safety-Utility Neuron Distribution}
\begin{figure}[h]
\centering
\includegraphics[width=\columnwidth]{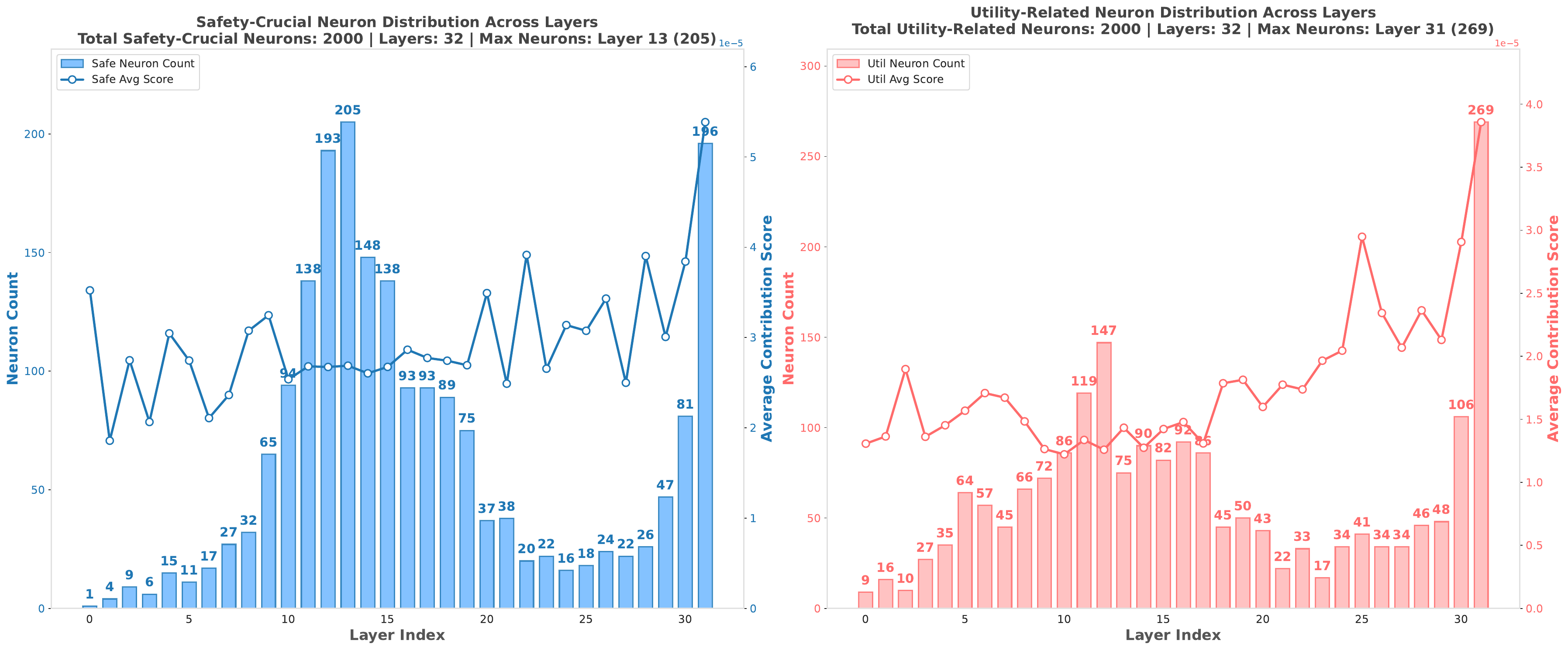}
\caption{Safety and utility neuron distribution across layers. The left plot illustrates safety neurons and the right plot illustrates utility neurons. Bars denote the count per layer, while lines indicate average contribution scores.}
\label{fig:safety_utility_neuron_dist}
\end{figure}

To gain a deeper understanding of how safety and utility capabilities are encoded within LLMs, we analyze the layer distribution of the identified neurons. This analysis provides insights into the architectural regions that are most critical for each aspect, further justifying our fine-grained intervention strategy. We visualize the distribution of 2000 safety neurons and 2000 utility neurons across the 32 layers of LLaMA3.1-8B-Instruct, along with their average contribution scores, as shown in Fig.~\ref{fig:safety_utility_neuron_dist}. 
Below are the observations from neuron distribution.

\textit{Distribution of Safety and Utility Neurons} Both safety-crucial and utility-related neurons are distributed across all layers, rather than being concentrated in a few specific layers. This observation supports that these capabilities are not localized but are complexly encoded throughout the network. It also explains why coarse-grained, layer-wise interventions inevitably lead to imbalance, as modifying an entire layer is likely to impact both safety and utility neurons residing within it.

\textit{Peak Concentrations in Different Layers}
Looking at the left of Fig.~\ref{fig:safety_utility_neuron_dist}, safety neurons show higher concentrations in middle layers, with a notable peak around Layer 13 and another significant concentration in later layers. Their average contribution scores also show fluctuations but remain relatively high across these layers. This suggests that safety-critical features are processed and refined in these intermediate and deeper layers.
In contrast, the right of Fig.~\ref{fig:safety_utility_neuron_dist} reveals that utility neurons tend to have higher concentrations in deeper layers, with a prominent peak in Layer 31. While present throughout, their density and average contribution scores generally increase towards the later layers. This aligns with the understanding that deeper layers in LLMs are often responsible for more abstract representations, complex reasoning, and factual knowledge, which are crucial for general utility.

\textit{Implications for Fine-Grained Modulation} 
The distribution patterns of safety and utility neurons underscore the necessity of a fine-grained intervention approach like NeuronTune. Since these crucial neurons are not perfectly segregated by layer, modification of an entire layer would inevitably impact both types of neurons. Our method, by pinpointing individual neurons and applying adaptive scaling factors, enables differentiated modulation for various neurons based on their specific functional roles. 
Besides, for a neuron $n_k$ that is in both $\mathcal{N}_S$ and $\mathcal{N}_U$, its scaling factor $\alpha_k$ undergoes dual optimizations during the inner loop: one steering $\alpha_k$ toward enhanced safety, while the other pushing it toward utility preservation. Crucially, the two competing objectives automatically drive the final optimized $\alpha_k$ toward an intermediate value, typically closer to the identity factor of 1. This automatic balancing act prevents over-emphasis on safety at the cost of utility degradation by preserving the neuron's dual roles, contributing significantly to the stability of the safety-utility trade-off.

We also observe 64--104 overlapping neurons under different selection thresholds, and the overlap increases as either the safety-critical or utility-related neuron set expands, confirming that the two functional roles are not fully separable.

This analysis provides empirical evidence supporting the architectural underpinnings of our NeuronTune and reinforces why fine-grained neuron modulation is a more effective strategy for balancing safety and utility in LLMs.

\section{Conclusion}
In this paper, we addressed the critical challenge of achieving a balance between robust safety and utility in LLMs. We first diagnose the inherent limitations of existing coarse-grained, layer-wise intervention strategies, which often struggle to simultaneously achieve robust safety and maintain high utility. To overcome it, we proposed \textbf{NeuronTune}, a novel fine-grained framework for safety-utility alignment.
NeuronTune is built upon two core steps: \textit{Pinpointing Neurons Responsible for Safety and Utility} and \textit{Editing Neurons via Adaptive Activation Adjustment}. This method allows for the targeted enhancement of safety-related activations and the preservation of utility-related knowledge. Furthermore, NeuronTune incorporates a tunable mechanism, offering flexible control over the number of modulated neurons to adapt to diverse application scenarios and varying demands for safety or utility.

\begin{credits}
\subsubsection{\ackname} This work was supported by Zhongguancun Academy (Grant No. 02012402)  and the National Natural Science Foundation of China (NSFC) (Grant No. 62576256).
\end{credits}

\bibliographystyle{splncs04}
\bibliography{custom}

\end{document}